\documentclass[journal]{IEEEtran}
\usepackage{cite}
\usepackage{amsmath,amssymb}
\usepackage{graphicx}
\usepackage{booktabs}
\usepackage{url}

\begin{document}

\title{Who Pays for Byzantine Robustness? Validity, Data-Share Dictatorships, and Fairness When Only a Few Hospitals Collaborate}

\author{Adnan Malkani\\
MSc Data Science, Sapienza University of Rome\\
\texttt{malkani.2245334@studenti.uniroma1.it}}

\maketitle

\begin{abstract}
Clinical federated learning (FL) collaborations typically connect only three to six hospitals, yet Byzantine-robust aggregation rules are designed and benchmarked on synthetic partitions across dozens of equal-sized clients. We conduct an empirical audit of eight standard aggregation rules: FedAvg, coordinate-wise median, trimmed mean, Krum, Multi-Krum, Bulyan, Weiszfeld geometric median (GeoMed), and Centered Clipping (CC). We evaluate both unweighted and sample-size-weighted variants across three realistic cross-silo medical datasets from the FLamby benchmark: Fed-Heart-Disease ($n = 4$ hospitals), Fed-TCGA-BRCA ($n = 6$ regions), and Fed-ISIC2019 ($n = 6$ centres). After reproducing published FLamby baselines and locking all hyperparameters via pre-specified (dated commits) validation rules, we execute 54,214 training runs (out of 84,344 result records; the remaining 30,130 configurations are invalid for the given number of clients and were not trained) across enumerated attacker coalitions ($f \in \{1, 2\}$) paired with exact-twin leave-out controls that decompose performance drops into honest data loss versus adversarial attack damage.

We report three findings. First, client-count validity checks fail to capture breakdown in skewed medical consortia: while $n = 6, f = 1$ satisfies every count condition except Bulyan's, assigning sample-size weights turns median and GeoMed into a clean-model dictatorship whenever a single hospital holds more than half the training data (such as centre 0 with 53.4\% of Fed-ISIC2019), and causes weighted median, GeoMed, and CC to collapse when that dominant centre attacks. Second, exact-twin leave-out controls reveal that when the largest hospital attacks, the apparent failure of count-based robust rules is predominantly caused by the loss of honest training data (7.5 to 10.4 percentage points on ISIC; 15.0 to 19.1 points on Heart) rather than adversarial perturbation (2.3 to 3.8 points on ISIC; $+0.3$ to $-2.6$ points on Heart). Conversely, weight-dependent rules suffer large true attack damage (16.1 to 25.1 points on ISIC). Third, unweighted robust rules consistently discount whichever institution generates the most atypical parameter updates, penalizing the smallest hospital on Heart (Switzerland, 6\% share) but penalizing larger institutions on TCGA and ISIC (median-type rules and CC: $\rho(\text{size}, \text{influence}) \in [-0.35, -0.77]$).
\end{abstract}

\section{Introduction}

Medical institutions rarely possess sufficient local patient volume to train generalizable diagnostic or prognostic models independently, while strict data-governance frameworks prohibit centralizing raw clinical records. Federated learning (FL) addresses this by keeping patient records on local hospital servers and aggregating only parameter updates at a central coordinator \cite{mcmahan2017communication,rieke2020future}. Unlike cross-device consumer applications involving thousands of mobile phones, real-world clinical consortia are small cross-silo networks: across the seven real-world healthcare datasets in the FLamby benchmark, each federation comprises between two and six institutions \cite{ogier2022flamby}.

Decentralized training, however, exposes the global model to corrupted updates. Hardware faults, miscalibrated imaging pipelines, label-mapping errors, or compromised hospital nodes can produce arbitrary parameter shifts, commonly modeled as Byzantine failures \cite{blanchard2017machine}. Because standard Federated Averaging (FedAvg) computes a linear combination of client updates, even a single corrupted institution can displace the global parameters arbitrarily far. To mitigate this risk, Byzantine-robust aggregation rules replace linear averaging with outlier-resistant operators, including coordinate-wise medians and trimmed means \cite{yin2018byzantine}, nearest-neighbour distance selection such as Krum and Bulyan \cite{blanchard2017machine,elmhamdi2018hidden}, Weiszfeld geometric medians \cite{pillutla2022robust}, and norm-bounded Centered Clipping \cite{karimireddy2021learning}.

Despite extensive theoretical and empirical study, standard evaluations of Byzantine-robust FL rely on 20 to 100 synthetic clients partitioned via Dirichlet distributions over benchmark datasets like MNIST or CIFAR-10 \cite{li2023experimental,li2024blades}. Such setups overlook two structural realities of clinical federations:
\begin{enumerate}
\item Distance-based selection rules require a minimum number of honest neighbours. Krum requires $n \geq 2f + 3$ clients to tolerate $f$ attackers and therefore cannot tolerate even a single faulty hospital in a four-centre consortium ($n = 4, f = 1$). Bulyan requires $n \geq 4f + 3$, rendering it mathematically undefined across all four- and six-hospital networks for any $f \geq 1$.
\item Patient populations vary by orders of magnitude across participating centres. In Fed-ISIC2019, a single dermatology centre holds 53.4\% of all training images (9,930 samples), whereas the smallest holds 1.9\% (351 samples). When robust aggregators incorporate sample-size weights to preserve clean accuracy, the breakdown point shifts from client count to data share.
\end{enumerate}

In this study, we evaluate how standard Byzantine-robust aggregation rules behave on natural hospital splits by addressing three research questions:
\begin{enumerate}
\item RQ-A: With four to six collaborating hospitals, which rules remain mathematically valid, which degenerate into simpler operators, and when does sample-size weighting hand unilateral control to a single institution?
\item RQ-B: When the central server defends against $f$ assumed attackers in clean and attacked regimes, what is the accuracy cost of robustness, and which specific hospitals bear that cost?
\item RQ-C: When a major clinical centre turns Byzantine and global performance drops sharply, how much of that degradation is true adversarial damage versus the unavoidable loss of that institution's honest training data?
\end{enumerate}

\textbf{Contributions.} We provide four contributions to the literature on trustworthy medical FL:
\begin{enumerate}
\item Alongside combinatorial client-count checks, we introduce a numerical weight-based validity check that identifies clean dictatorships (where one hospital holding $> 50\%$ of sample weight completely determines the output of weighted median and GeoMed) and weight-driven breakdowns.
\item We pair every attacked configuration with a deterministic leave-out control ($n - 1$ clients for $f = 1$; $n - 2$ for $f = 2$) in which the excluded hospital still advances its random batch generator without training. This isolates honest data loss from true adversarial attack damage.
\item We show that unweighted median, trimmed mean, GeoMed and CC discount whichever hospital exhibits the largest update distance from the consensus, penalizing the smallest hospital on Fed-Heart-Disease, but penalizing larger, more data-rich institutions on Fed-TCGA-BRCA and Fed-ISIC2019.
\item We release our complete PyTorch benchmark (byzantine-fl-bench)\footnote{\url{https://github.com/adnanmalkani/byzantine-fl-bench}}, pre-specified hyperparameter selection protocol, FLamby baseline reproductions, and all 84,344 experimental run records.
\end{enumerate}

\section{Background and related work}

\subsection{Byzantine-robust aggregation rules}

Let $\Delta_i^{(t)} = w_i^{(t)} - w^{(t-1)}$ denote the local model update transmitted by hospital $i \in \{1, \ldots, n\}$ at communication round $t$, trained on $m_i$ local patient records with relative weight $p_i = m_i / \sum_{j=1}^{n} m_j$. Standard FedAvg updates the global model via $\sum_{i=1}^{n} p_i \Delta_i^{(t)}$ (or $\frac{1}{n} \sum_{i=1}^{n} \Delta_i^{(t)}$ when unweighted) \cite{mcmahan2017communication}.

To bound the influence of $f$ Byzantine clients, Coordinate-wise Median (\texttt{median}) computes the one-dimensional median independently along each parameter coordinate, while Trimmed Mean (\texttt{trimmed\_mean}) discards the largest $f$ and smallest $f$ values per coordinate before averaging the remaining $n - 2f$ entries \cite{yin2018byzantine}. Krum scores each update by the sum of squared Euclidean distances to its $n - f - 2$ nearest neighbours and outputs the single update with the lowest score; Multi-Krum (\texttt{multi\_krum}) averages the $m$ lowest-scoring updates, where $m$ is a free parameter ($1 \leq m \leq n$); following the original experiments \cite{blanchard2017machine}, we set $m = n - f$. Bulyan (\texttt{bulyan}) iteratively selects $n - 2f$ candidates via Krum and applies a coordinate-wise trimmed mean removing $2f$ extremes, requiring $n \geq 4f + 3$ \cite{elmhamdi2018hidden}. Geometric Median (\texttt{geomed}), the core primitive of Robust Federated Averaging (RFA), minimizes the sum of Euclidean distances $\sum_{i=1}^{n} \alpha_i \|z - \Delta_i^{(t)}\|_2$ via smoothed Weiszfeld iterations \cite{pillutla2022robust}. Finally, Centered Clipping (\texttt{cc}) iteratively clips client updates toward a reference point $v$ within radius $\tau$: $v \leftarrow v + \sum_{i=1}^{n} \alpha_i (\Delta_i^{(t)} - v) \min(1, \tau / \|\Delta_i^{(t)} - v\|_2)$ \cite{karimireddy2021learning}.

\subsection{Adversarial threat vectors}

Evaluations of Byzantine resilience span both untargeted untrustworthy updates and worst-case omniscient perturbations. Label Flipping (\texttt{label\_flip}) corrupts local training targets prior to gradient computation (inverting binary outcomes or shifting multiclass labels via $y \mapsto (C - 1) - y$). Sign Flipping (\texttt{sign\_flip}) transmits $-\Delta_i^{(t)}$, while Gaussian Noise (\texttt{gaussian}) replaces the update with isotropic noise $\mathcal{N}(0, \sigma^2 I)$. Other model-poisoning attacks exploit knowledge of the honest client distribution: Inner Product Manipulation (\texttt{ipm}) transmits $-\epsilon \bar{\Delta}_{\text{honest}}$ ($\epsilon = 0.5$) to reverse the descent direction without triggering Euclidean norm filters \cite{xie2020fall}, whereas A Little Is Enough (\texttt{alie}) sets each coordinate to $\mu_{\text{honest}} - z \cdot \sigma_{\text{honest}}$ so that malicious updates hide within the natural variance of honest hospitals \cite{baruch2019little}.

\subsection{Heterogeneity, small client counts, and clinical benchmarks}

Recent theoretical analyses demonstrate that inter-client data heterogeneity degrades Byzantine resilience. Allouah et al.\ \cite{allouah2023robust} establish tight error bounds showing that under non-IID distributions, the effective breakdown point of robust rules drops well below $1/2$. Karimireddy et al.\ \cite{karimireddy2022byzantine} show that coordinate-wise median rules consistently bias models toward homogeneous subsets of clients and propose random bucketing prior to aggregation; however, bucketing requires partitioning clients into groups of size $s \geq 2$ while maintaining $n/s > 2f$, which is impossible at $n = 4$ and at $n = 6$ with $f = 2$; at $n = 6$, $f = 1$ it leaves only three buckets.

While general FL security suites such as Blades \cite{li2024blades} and Li et al.\ \cite{li2023experimental} benchmark defences on synthetic partitions, and FLamby \cite{ogier2022flamby} establishes natural clinical splits without Byzantine attacks, very few studies intersect the two. Touat and Bouchenak \cite{touat2023towards} observe demographic unfairness under robust aggregation on tabular benchmarks, and FedSECA \cite{benjamin2025fedseca} evaluates sign-election aggregation on Fed-ISIC2019. To our knowledge, no prior work audits the validity breakdown of weighted vs.\ unweighted rules across multiple FLamby datasets, enumerates all hospital coalitions, or isolates honest data loss from adversarial attack damage via leave-out controls.

\section{Experimental protocol and pre-specified design}

\subsection{Clinical datasets and baseline reproductions}

We evaluate three natural cross-silo datasets from FLamby \cite{ogier2022flamby} alongside a 20-client Dirichlet-split ($\alpha = 0.5$) MNIST sanity benchmark. Table~\ref{tab:datasets} details the institutional splits, sample shares, models, and pre-specified (dated commits) training parameters.

\begin{table*}[!t]
\centering
\caption{Dataset characteristics, natural hospital training/test distributions (ordered by training share), and pre-specified plateau hyperparameters.}
\label{tab:datasets}
\resizebox{\textwidth}{!}{
\begin{tabular}{@{}lrp{5.2cm}p{2.4cm}p{2.2cm}rrrr@{}}
\toprule
Dataset & $n$ & Client: training share (train/test size), largest first & Model & Main metric & lr & Rounds & $\tau$ (CC) & Seeds \\
\midrule
Fed-Heart-Disease & 4 & Cleveland 40.9\% (199/104), Hungary 35.4\% (172/89), VA 17.5\% (85/45), Switzerland 6.2\% (30/16) & logistic regression & accuracy (test-size weighted) & 0.001 & 106 & 0.03 & 20 \\
Fed-TCGA-BRCA & 6 & Northeast 28.6\% (248/63), West 18.9\% (164/42), South 18.0\% (156/40), Midwest 14.9\% (129/33), Europe 14.9\% (129/33), Canada 4.6\% (40/11) & linear Cox model & C-index (stratified) & 0.003 & 243 & 0.03 & 20 ($f=2$: 5) \\
Fed-ISIC2019 & 6 & Centre 0 53.4\% (9930/2483), Centre 1 17.0\% (3163/791), Centre 2 14.5\% (2691/672), Centre 3 9.7\% (1807/452), Centre 4 3.5\% (655/164), Centre 5 1.9\% (351/88) & linear head on frozen EfficientNet-B0 features & balanced accuracy (pooled) & 0.01 & 101 & 1 & 20 \\
\bottomrule
\end{tabular}
}
\end{table*}

Before modifying any training horizons, we verified our pipeline against FLamby's published FedAvg baselines using FLamby's exact hyperparameters (Table~\ref{tab:repro}; Fig.~\ref{fig:repro} in the Appendix). On Fed-Heart-Disease (15 rounds, lr 0.001), our implementation reproduces FLamby's pooled test accuracy of $74.8 \pm 2.7\%$ seed for seed across all 5 benchmark seeds. On Fed-TCGA-BRCA (5 rounds, lr 0.1), our pooled C-index of $0.710 \pm 0.196$ matches FLamby's $0.735 \pm 0.175$ within 0.14 standard deviations; inspection of per-round trajectories confirmed identical weight initializations, with minor batch-order differences determining which individual seed collapses at lr $= 0.1$. On Fed-ISIC2019, executing FLamby's full EfficientNet-B0 fine-tuning code across seeds 42 and 43 yields a pooled balanced accuracy of 61.5\% (63.9\% and 59.1\%), well inside the pre-specified acceptance interval of $53.4 \pm 11.4\%$ ([42.0\%, 64.8\%]).

\begin{table}[!t]
\centering
\caption{Reproduction of published FLamby FedAvg baselines using FLamby's original hyperparameters.}
\label{tab:repro}
\resizebox{\columnwidth}{!}{
\begin{tabular}{@{}l l l r c@{}}
\toprule
Test set & Ours & FLamby (published) & Difference & Within 1 s.d. \\
\midrule
\multicolumn{5}{@{}l}{\textit{Fed-Heart-Disease, accuracy, 5 seeds}} \\
Pooled & 0.748 $\pm$ 0.027 & 0.748 $\pm$ 0.027 & +0.000 & yes \\
Cleveland & 0.727 $\pm$ 0.019 & 0.731 $\pm$ 0.022 & $-$0.004 & yes \\
Hungary & 0.717 $\pm$ 0.037 & 0.719 $\pm$ 0.033 & $-$0.002 & yes \\
VA & 0.640 $\pm$ 0.051 & 0.640 $\pm$ 0.051 & +0.000 & yes \\
Switzerland & 0.650 $\pm$ 0.114 & 0.650 $\pm$ 0.114 & +0.000 & yes \\
\midrule
\multicolumn{5}{@{}l}{\textit{Fed-TCGA-BRCA, C-index, 5 seeds}} \\
Pooled & 0.710 $\pm$ 0.196 & 0.735 $\pm$ 0.175 & $-$0.025 & yes \\
Northeast & 0.698 $\pm$ 0.134 & 0.745 $\pm$ 0.099 & $-$0.047 & yes \\
West & 0.663 $\pm$ 0.211 & 0.691 $\pm$ 0.190 & $-$0.028 & yes \\
South & 0.664 $\pm$ 0.292 & 0.627 $\pm$ 0.228 & +0.036 & yes \\
Midwest & 0.617 $\pm$ 0.089 & 0.675 $\pm$ 0.120 & $-$0.058 & yes \\
Europe & 0.796 $\pm$ 0.421 & 0.787 $\pm$ 0.373 & +0.009 & yes \\
Canada & 0.800 $\pm$ 0.298 & 0.867 $\pm$ 0.298 & $-$0.067 & yes \\
\midrule
\multicolumn{5}{@{}l}{\textit{Fed-ISIC2019, balanced accuracy, FLamby's code, 2 seeds (ours) vs 5}} \\
Pooled & 0.615 (0.639, 0.591) & 0.534 $\pm$ 0.114 & +0.081 & yes \\
Centre 0 & 0.650 (0.668, 0.632) & 0.560 $\pm$ 0.136 & +0.090 & yes \\
Centre 1 & 0.601 (0.605, 0.596) & 0.606 $\pm$ 0.110 & $-$0.005 & yes \\
Centre 2 & 0.721 (0.722, 0.720) & 0.645 $\pm$ 0.136 & +0.076 & yes \\
Centre 3 & 0.488 (0.544, 0.431) & 0.423 $\pm$ 0.091 & +0.065 & yes \\
Centre 4 & 0.396 (0.386, 0.405) & 0.387 $\pm$ 0.049 & +0.009 & yes \\
Centre 5 & 0.660 (0.717, 0.602) & 0.612 $\pm$ 0.088 & +0.047 & yes \\
\bottomrule
\end{tabular}
}
\end{table}

\subsection{Pre-specified plateau rules and design amendments}

At FLamby's default 15 rounds on Heart and 9 rounds on ISIC, FedAvg remains under-converged, which would confound differences in convergence speed with Byzantine robustness. Furthermore, FLamby's lr $= 0.1$ on TCGA causes 50\% of clean FedAvg seeds (10/20) to collapse on validation data. We therefore fixed pre-specified validation rules (using a stratified 10\% holdout of training data, never touching test sets) prior to launching main grids:
\begin{enumerate}
\item On TCGA and ISIC, we screened lr $\in \{0.1, 0.03, 0.01, 0.003\}$ (TCGA) and $\{0.3, 0.1, 0.03, 0.01, 0.003\}$ (ISIC) over 20 seeds for 200 rounds, selecting the largest rate with zero collapsed seeds. On TCGA, two hard validation splits (seeds 13 and 19) failed to reach the 0.60 threshold at any learning rate; Amendment 1 (dated 12 Oct 2026) excluded rate-invariant hard splits from the stop trigger and extended screening to 400 rounds, selecting lr $= 0.003$. On ISIC, we extracted 1280-dimensional features from a frozen ImageNet EfficientNet-B0 backbone \cite{tan2019efficientnet} (centre-cropped at $200 \times 200$ px) so all rules compare on identical representations using SGD (avoiding Adam's coordinate-wise rescaling). Because no learning rate collapsed on ISIC, the largest-stable-rate rule trivially picked 0.3, which failed to plateau; Amendment 2 (dated 17 Oct 2026) selected FLamby's published FedAvg rate of lr $= 0.01$.
\item The stopping horizon $R^*$ is defined as the earliest round after which mean validation performance over 20 seeds never dips more than 0.5 points below the late-round plateau average (151--200, or 301--400 for TCGA). This fixed $R^* = 106$ for Heart, $R^* = 243$ for TCGA, and $R^* = 101$ for ISIC.
\item The Centered Clipping radius $\tau$ follows Rule v2.1: $\tau_0$ is initialized as the median Euclidean distance of honest updates from the previous aggregate and snapped to a logarithmic grid, overridden only if an alternative $\tau$ improves both weighted and unweighted validation accuracy with a 95\% CI lower bound exceeding $+0.5$ points. This selected $\tau = 0.03$ on Heart, $\tau = 0.03$ on TCGA, and $\tau = 1.0$ on ISIC.
\end{enumerate}

\subsection{Weighting separation, enumerated attackers, and metrics}

To isolate the effect of institutional size from outlier suppression (RQ-B), every rule is evaluated in an unweighted variant and a sample-size-weighted variant (w). In weighted variants, sample weights $p_i$ enter only the final aggregation operator (weighted coordinate-wise median, weighted Weiszfeld iterations for GeoMed, weighted clipping average for CC, and weighted averaging of kept candidates for Trimmed Mean, Multi-Krum, and Bulyan); candidate selection and trimming counts remain strictly unweighted. Krum selects a single client, so its two variants are identical.

Because client sizes vary, random attacker sampling buries hospital-identity effects inside seed variance. We therefore enumerate every single hospital as the attacker for $f = 1$ (4 on Heart, 6 on TCGA and ISIC) and every hospital pair for $f = 2$ ($\binom{4}{2} = 6$ on Heart, $\binom{6}{2} = 15$ on TCGA and ISIC). In clean baselines ($f_{\text{actual}} = 0$), the server still defends against $f_{\text{assumed}} \in \{1, 2\}$ so that clean rows measure the true operational cost of turning on a defence.

Alongside Baruch's standard ALIE formula \cite{baruch2019little}, we pre-specified a fixed-radius variant, \texttt{alie\_z1} ($z = 1.0$). At $n \in \{4, 6\}$ and $f = 1$, Baruch's normal-CDF calculation yields $z = 0$, causing the attacker to transmit the exact mean of all updates (a degenerate non-attack). We flag $z = 0$ runs as degenerate and exclude them from the ``mean of real attacks.''

\textbf{Evaluation metrics:} Every run is scored by the mean of its final 3 evaluations.
\begin{enumerate}
\item Fed-Heart-Disease uses the patient-weighted average of per-hospital test accuracies (each hospital normalized by its local training statistics), plus the accuracy of the worst honest hospital.
\item Fed-TCGA-BRCA uses the stratified C-index over all 608 within-region comparable patient pairs pooled (preventing Canada's 3 pairs from carrying disproportionate weight). For the worst honest region and regional gap, only regions with $\geq 3$ test deaths qualify (Northeast: 14, West: 8, South: 4, Midwest: 3; excluding Europe: 2 and Canada: 1).
\item Fed-ISIC2019 uses pooled balanced accuracy across all 4,650 test images. Because several centres lack rare lesion classes entirely, the worst-honest-centre metric evaluates balanced accuracy over classes with at least $k = 5$ test images at that centre ($k = 10$ sensitivity check in the Appendix, Table~\ref{tab:k10}). We also identified 49 byte-identical duplicate image pairs in FLamby's ISIC split, 17 of which straddle the train/test boundary (16 at C0, 1 at C5); a dedicated sensitivity sweep removing those 17 test images confirms all clean metrics shift by $\leq 0.16$ points.
\end{enumerate}

\subsection{Exact-twin leave-out controls}
\label{sec:loo}

When a major centre (e.g., Cleveland at 41\% or ISIC C0 at 53.4\%) turns Byzantine, performance drops reflect both the loss of its honest data and the damage inflicted by its poisoned updates. We separate these effects by running clean leave-one-out ($n - 1$ clients, $f_{\text{assumed}} = 1$) and leave-two-out ($n - 2$ clients, $f_{\text{assumed}} = 2$) controls for every hospital and pair. To ensure controls are exact twins of the clean and attacked runs, excluded hospitals still draw their batch indices from the shared pseudorandom stream each round without training, guaranteeing that all remaining hospitals process bit-identical mini-batches across clean, control, and attacked executions, except under the Gaussian attack, whose noise uses the same random stream. For each seed $s$, total degradation decomposes additively over the same honest client set:
\begin{equation}
\begin{aligned}
&\underbrace{M_{\text{attacked}}(s) - M_{\text{clean}}(s)}_{\text{Total Drop}}\\
&\quad= \underbrace{M_{\text{control}}(s) - M_{\text{clean}}(s)}_{\text{Data Loss (Client Absent)}}\\
&\quad\quad+ \underbrace{M_{\text{attacked}}(s) - M_{\text{control}}(s)}_{\text{True Attack Damage}}.
\end{aligned}
\label{eq:decomp}
\end{equation}

In our main leave-out tables, attack damage is averaged across all non-degenerate attacks excluding ALIE: because our ALIE vector is constructed from the mean of all benign updates including the attacker's own honest update, comparing an ALIE attacker against a leave-out control partially returns the missing hospital's honest gradient, artificially inflating attack damage by up to $+24.9$ points.

\section{Results}

\subsection{RQ-A: Validity must be judged by data share, not client count}

\begin{figure*}[!t]
\centering
\includegraphics[width=0.66\textwidth]{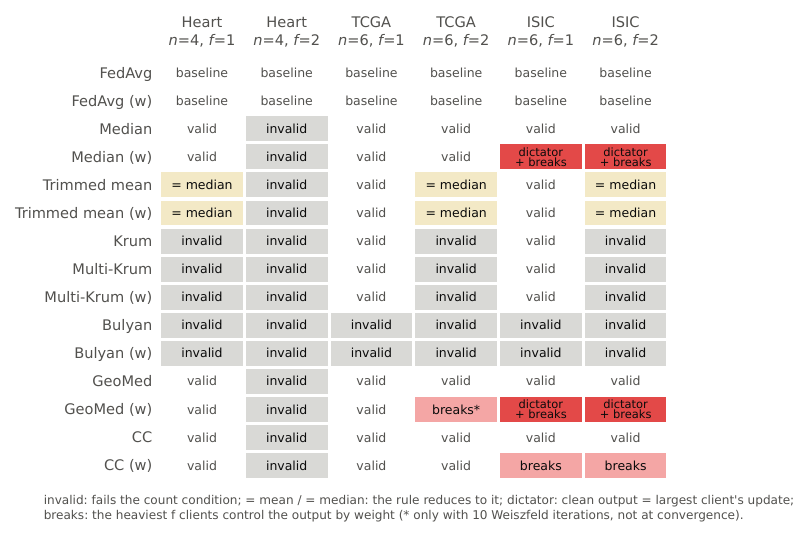}
\caption{Status of every rule variant in each federation $(n, f)$: invalid by client count, reducing to the mean or median, or flagged by data share as a dictator (the clean output is the largest client's update) or as breaking when the heaviest $f$ clients attack.}
\label{fig:validity}
\end{figure*}

Table~\ref{tab:validity} and Fig.~\ref{fig:validity} summarize the combinatorial and weight-based validity envelope across all three datasets. Under standard client-count conditions, $n = 4, f = 1$ (Heart) immediately disqualifies Krum, Multi-Krum ($n \geq 2f + 3 = 5$), and Bulyan ($n \geq 4f + 3 = 7$), while reducing unweighted Trimmed Mean identically to Coordinate-wise Median (verified across 500/500 runs). At $n = 4, f = 2$, no Byzantine-robust rule is mathematically valid. At $n = 6$ (TCGA and ISIC), Bulyan remains invalid at $f = 1$, and both Krum and Multi-Krum fail at $f = 2$, with unweighted Trimmed Mean again collapsing into Median.

\begin{table}[!t]
\centering
\caption{Combinatorial client-count validity and weight-based breakdown envelope across natural hospital splits. $\checkmark$: valid; $\times$: invalid by client count; = median: the rule reduces to the median. Weighted columns (heaviest client's share in brackets): D = dictator when clean; B1/B2 = the heaviest 1/2 clients break the rule; $^*$ = only with 10 Weiszfeld iterations; n/a = invalid by count.}
\label{tab:validity}
\resizebox{\columnwidth}{!}{
\begin{tabular}{@{}l cccc ccc@{}}
\toprule
& \multicolumn{4}{c}{By client count} & \multicolumn{3}{c}{Weighted, by data share} \\
\cmidrule(lr){2-5}\cmidrule(l){6-8}
& \multicolumn{2}{c}{$n=4$ (Heart)} & \multicolumn{2}{c}{$n=6$ (TCGA, ISIC)} & Heart & TCGA & ISIC \\
Rule & $f=1$ & $f=2$ & $f=1$ & $f=2$ & (40.9\%) & (28.6\%) & (53.4\%) \\
\midrule
FedAvg & -- & -- & -- & -- & -- & -- & -- \\
Median & $\checkmark$ & $\times$ & $\checkmark$ & $\checkmark$ & $\checkmark$ & $\checkmark$ & D, B1, B2 \\
Trimmed mean & = median & $\times$ & $\checkmark$ & = median & $\checkmark$ & $\checkmark$ & $\checkmark$ \\
Krum & $\times$ & $\times$ & $\checkmark$ & $\times$ & ignores $w$ & ignores $w$ & ignores $w$ \\
Multi-Krum & $\times$ & $\times$ & $\checkmark$ & $\times$ & n/a & $\checkmark$ & $\checkmark$ \\
Bulyan & $\times$ & $\times$ & $\times$ & $\times$ & n/a & n/a & n/a \\
GeoMed & $\checkmark$ & $\times$ & $\checkmark$ & $\checkmark$ & $\checkmark$ & B2$^*$ & D, B1, B2 \\
CC & $\checkmark$ & $\times$ & $\checkmark$ & $\checkmark$ & $\checkmark$ & $\checkmark$ & B1, B2 \\
\bottomrule
\end{tabular}
}
\end{table}

Client counts miss weight-induced breakdowns. On Fed-ISIC2019 ($n = 6, f = 1$), every rule except Bulyan passes count checks. However, because centre 0 holds 53.4\% of the training data, any weighted median operator reaches the 50\% cumulative weight threshold inside centre 0's update. Consequently, in clean training ($f = 0$), median (w) and geomed (w) act as single-hospital dictatorships: centre 0 receives a relative influence of $1.87 = 1/0.534$ (100\% of the aggregate), while centres 1 through 5 receive 0.00 to 0.01 relative influence. When centre 0 attacks, median (w) and geomed (w) adopt the malicious update entirely, and cc (w) experiences large drift because the 53.4\% attacker out-pulls the 46.6\% honest coalition on every clipping step. A similar dictatorship emerges in Heart leave-one-out controls when Cleveland or Hungary is held out, leaving the remaining major hospital with 59.9\% or 63.4\% of total weight.

\subsection{RQ-B: Clean cost of robustness and who pays}

Table~\ref{tab:clean} reports clean performance when the server defends against $f_{\text{assumed}} = 1$ attacker. Sample-size weighting improves clean average performance across all three datasets when all hospitals are honest (FedAvg gains $+3.4$ points on Heart, $+0.6$ on TCGA, and $+5.2$ on ISIC), likely because larger centres provide lower-variance updates. However, weighting can penalize smaller hospitals: on Heart, switching from unweighted to weighted FedAvg raises the average from 70.9\% to 74.3\% but drops the worst-served hospital from 66.6\% (Hungary) to 63.0\% (VA). Under median (w), Switzerland (6.2\% share) receives exactly 0.00 coordinate selection share, driving the worst hospital down to 58.8\%. On ISIC, the clean dictatorship of median (w) and geomed (w) reduces centre 1's local accuracy from 74.3\% (under weighted FedAvg) to 43.1\% and centre 4's from 48.8\% to 33.5\%.

\begin{table*}[!t]
\centering
\caption{Clean performance ($f_{\text{actual}} = 0$, $f_{\text{assumed}} = 1$, 20 seeds): average and worst-hospital metric (mean $\pm$ 95\% CI over seeds), the least-influential client with its relative influence, and Spearman rank correlation $\rho(\text{size}, \text{influence})$, for unweighted and weighted (w) rules.}
\label{tab:clean}
\resizebox{\textwidth}{!}{
\begin{tabular}{@{}l cccc cccc@{}}
\toprule
& \multicolumn{4}{c}{Unweighted} & \multicolumn{4}{c}{Weighted (w)} \\
\cmidrule(lr){2-5}\cmidrule(l){6-9}
Rule & Avg & Worst & Least influence & $\rho$ & Avg & Worst & Least influence & $\rho$ \\
\midrule
\multicolumn{9}{@{}l}{\textit{Fed-Heart-Disease (accuracy)}} \\
FedAvg & 70.9{\scriptsize\,$\pm$0.4} & 66.6{\scriptsize\,$\pm$0.7} & -- & -- & 74.3{\scriptsize\,$\pm$0.3} & 63.0{\scriptsize\,$\pm$0.7} & -- & -- \\
Median & 72.8{\scriptsize\,$\pm$0.3} & 69.8{\scriptsize\,$\pm$0.6} & Switzerland (0.49) & 0.20 & 73.8{\scriptsize\,$\pm$0.3} & 58.8{\scriptsize\,$\pm$0.6} & Switzerland (0.00) & 0.20 \\
Trimmed mean & 72.8{\scriptsize\,$\pm$0.3} & 69.8{\scriptsize\,$\pm$0.6} & Switzerland (0.49) & 0.20 & 74.3{\scriptsize\,$\pm$0.3} & 65.4{\scriptsize\,$\pm$0.9} & Switzerland (0.53) & 0.20 \\
Krum & invalid &  &  &  & \multicolumn{4}{c}{= Krum (ignores weights)} \\
Multi-Krum & invalid &  &  &  & invalid &  &  &  \\
Bulyan & invalid &  &  &  & invalid &  &  &  \\
GeoMed & 70.1{\scriptsize\,$\pm$0.5} & 65.8{\scriptsize\,$\pm$0.9} & Switzerland (0.66) & 0.40 & 74.5{\scriptsize\,$\pm$0.4} & 63.4{\scriptsize\,$\pm$0.8} & Switzerland (0.13) & 1.00 \\
CC & 73.4{\scriptsize\,$\pm$0.4} & 70.7{\scriptsize\,$\pm$0.6} & Switzerland (0.64) & 0.40 & 74.7{\scriptsize\,$\pm$0.3} & 61.6{\scriptsize\,$\pm$0.7} & Switzerland (0.49) & 1.00 \\
\midrule
\multicolumn{9}{@{}l}{\textit{Fed-TCGA-BRCA (C-index)}} \\
FedAvg & 82.0{\scriptsize\,$\pm$0.4} & 64.3{\scriptsize\,$\pm$0.9} & -- & -- & 82.6{\scriptsize\,$\pm$0.3} & 63.3{\scriptsize\,$\pm$0.7} & -- & -- \\
Median & 79.2{\scriptsize\,$\pm$0.6} & 65.6{\scriptsize\,$\pm$1.4} & South (0.64) & $-$0.35 & 80.5{\scriptsize\,$\pm$0.4} & 67.0{\scriptsize\,$\pm$1.5} & Europe (0.69) & $-$0.38 \\
Trimmed mean & 80.5{\scriptsize\,$\pm$0.4} & 66.3{\scriptsize\,$\pm$1.2} & South (0.80) & $-$0.55 & 80.7{\scriptsize\,$\pm$0.5} & 66.2{\scriptsize\,$\pm$0.7} & South (0.85) & $-$0.55 \\
Krum & 69.0{\scriptsize\,$\pm$1.9} & 51.1{\scriptsize\,$\pm$2.6} & South (0.00) & $-$0.75 & \multicolumn{4}{c}{= Krum (ignores weights)} \\
Multi-Krum & 81.2{\scriptsize\,$\pm$0.3} & 63.8{\scriptsize\,$\pm$1.1} & West (0.58) & $-$0.52 & 81.8{\scriptsize\,$\pm$0.3} & 63.3{\scriptsize\,$\pm$1.1} & West (0.59) & $-$0.52 \\
Bulyan & invalid &  &  &  & invalid &  &  &  \\
GeoMed & 79.7{\scriptsize\,$\pm$0.6} & 64.4{\scriptsize\,$\pm$1.5} & West (0.40) & $-$0.61 & 81.1{\scriptsize\,$\pm$0.3} & 66.2{\scriptsize\,$\pm$1.0} & West (0.50) & $-$0.41 \\
CC & 81.9{\scriptsize\,$\pm$0.3} & 64.3{\scriptsize\,$\pm$1.0} & West (0.87) & $-$0.52 & 82.5{\scriptsize\,$\pm$0.3} & 63.3{\scriptsize\,$\pm$0.8} & West (0.86) & $-$0.52 \\
\midrule
\multicolumn{9}{@{}l}{\textit{Fed-ISIC2019 (balanced accuracy)}} \\
FedAvg & 49.6{\scriptsize\,$\pm$0.2} & 43.8{\scriptsize\,$\pm$0.3} & -- & -- & 54.8{\scriptsize\,$\pm$0.3} & 48.8{\scriptsize\,$\pm$0.4} & -- & -- \\
Median & 44.3{\scriptsize\,$\pm$0.1} & 37.1{\scriptsize\,$\pm$0.2} & c0 (0.48) & $-$0.77 & 48.9{\scriptsize\,$\pm$0.3} & 33.5{\scriptsize\,$\pm$0.9} & c1 (0.00) & 0.66 \\
Trimmed mean & 46.5{\scriptsize\,$\pm$0.1} & 39.5{\scriptsize\,$\pm$0.2} & c0 (0.63) & $-$0.77 & 48.8{\scriptsize\,$\pm$0.2} & 42.4{\scriptsize\,$\pm$0.3} & c0 (0.78) & $-$0.60 \\
Krum & 23.2{\scriptsize\,$\pm$0.7} & 17.7{\scriptsize\,$\pm$0.1} & c0 (0.00) & $-$0.07 & \multicolumn{4}{c}{= Krum (ignores weights)} \\
Multi-Krum & 48.0{\scriptsize\,$\pm$0.2} & 36.8{\scriptsize\,$\pm$0.3} & c3 (0.00) & 0.03 & 52.7{\scriptsize\,$\pm$0.2} & 39.8{\scriptsize\,$\pm$0.6} & c3 (0.00) & 0.03 \\
Bulyan & invalid &  &  &  & invalid &  &  &  \\
GeoMed & 44.2{\scriptsize\,$\pm$0.2} & 37.8{\scriptsize\,$\pm$0.2} & c0 (0.43) & $-$0.60 & 49.0{\scriptsize\,$\pm$0.3} & 33.6{\scriptsize\,$\pm$0.9} & c3 (0.00) & 0.09 \\
CC & 49.4{\scriptsize\,$\pm$0.2} & 43.5{\scriptsize\,$\pm$0.3} & c0 (0.83) & $-$0.71 & 54.6{\scriptsize\,$\pm$0.3} & 48.8{\scriptsize\,$\pm$0.3} & c3 (0.59) & 0.43 \\
\bottomrule
\end{tabular}
}
\end{table*}

\begin{figure*}[!t]
\centering
\includegraphics[width=\textwidth]{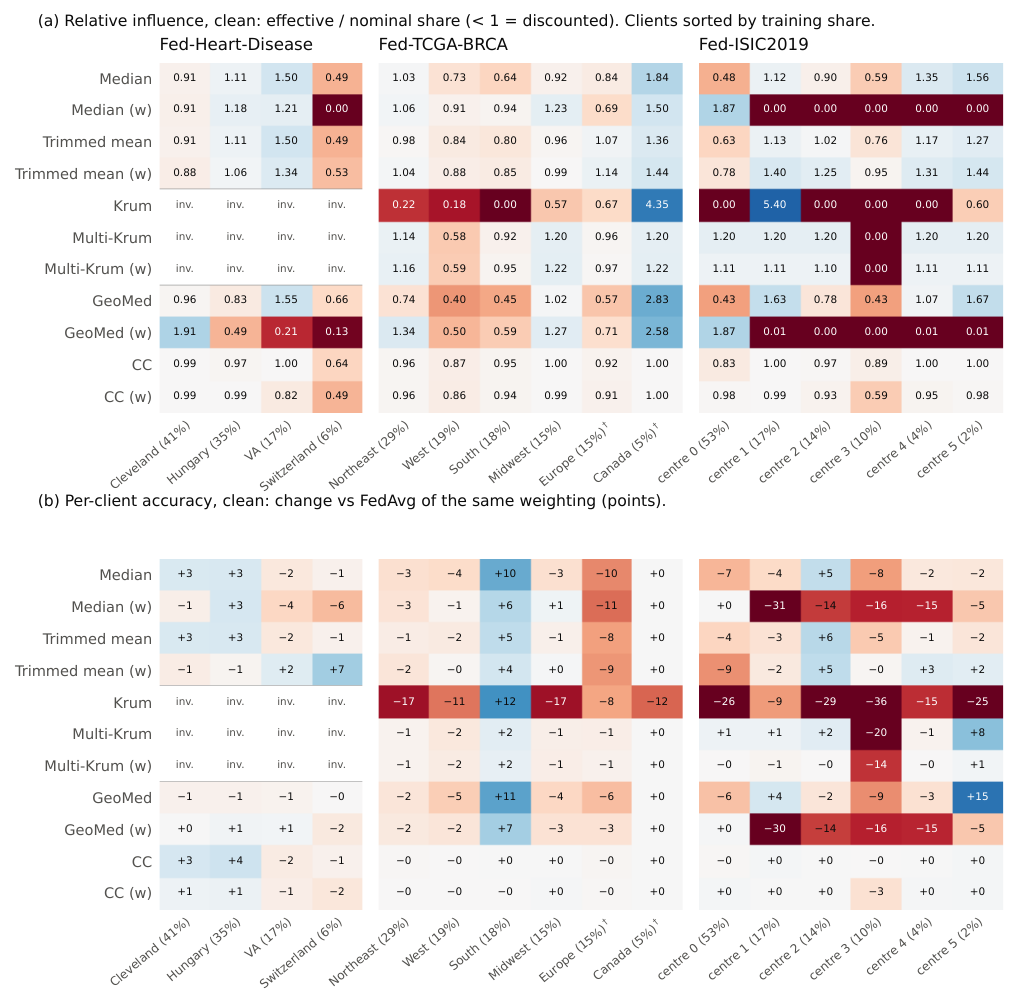}
\caption{Who pays in clean runs ($f_{\text{assumed}} = 1$): (a) each client's relative influence per rule (effective over nominal share, $< 1$ = discounted; clients sorted by training share) and (b) each client's accuracy change against FedAvg of the same weighting.}
\label{fig:who_pays}
\end{figure*}

Auditing relative influence across hospitals (Fig.~\ref{fig:who_pays}) reveals a pattern: unweighted median, trimmed mean, GeoMed and CC allocate influence inversely to update distance ($\rho(\text{distance}, \text{influence}) \in [-0.49, -1.00]$ on all three datasets; on Fed-Heart-Disease each $\rho$ is computed over only four hospitals), not hospital size. Because update distance depends on local data distribution and step norms, the hospital discounted by unweighted robust rules flips across clinical tasks:
\begin{enumerate}
\item On Fed-Heart-Disease, the smallest hospital (Switzerland, 30 patients) generates the most distant updates and is discounted by every robust rule (0.49--0.66 relative influence unweighted; $\rho(\text{size}, \text{influence}) > 0$).
\item On Fed-TCGA-BRCA and Fed-ISIC2019, larger institutions exhibit larger update distances while smaller centres sit closer to the aggregate. Consequently, unweighted median, trimmed mean, GeoMed and CC down-weight larger hospitals ($\rho(\text{size}, \text{influence}) \in [-0.35, -0.61]$ on TCGA and $[-0.60, -0.77]$ on ISIC); Krum and Multi-Krum, which select whole updates, are reported separately in Table~\ref{tab:clean}. On ISIC, unweighted median and GeoMed reduce centre 0's influence to 0.48 and 0.43 of its nominal 1/6 share while boosting the smallest centre (C5) to 1.56--1.67$\times$ its share (raising C5's local accuracy from 76.6\% under FedAvg to 91.3\% under GeoMed). On TCGA, Krum selects the smallest region (Canada, 40 patients) at 4.4$\times$ its nominal share, causing clean Krum to drop 13 points below FedAvg (69.0 vs.\ 82.0).
\end{enumerate}

Across all three datasets, unweighted Centered Clipping (\texttt{cc}) incurs virtually zero clean robustness penalty: it remains within 0.1--0.2 points of unweighted FedAvg on TCGA and ISIC, and outperforms unweighted FedAvg by $+2.5$ points on Heart (73.4\% vs.\ 70.9\%).

\subsection{Single-attacker resilience ($f = 1$) and stealthy in-distribution harm}

\begin{figure*}[!t]
\centering
\includegraphics[width=\textwidth]{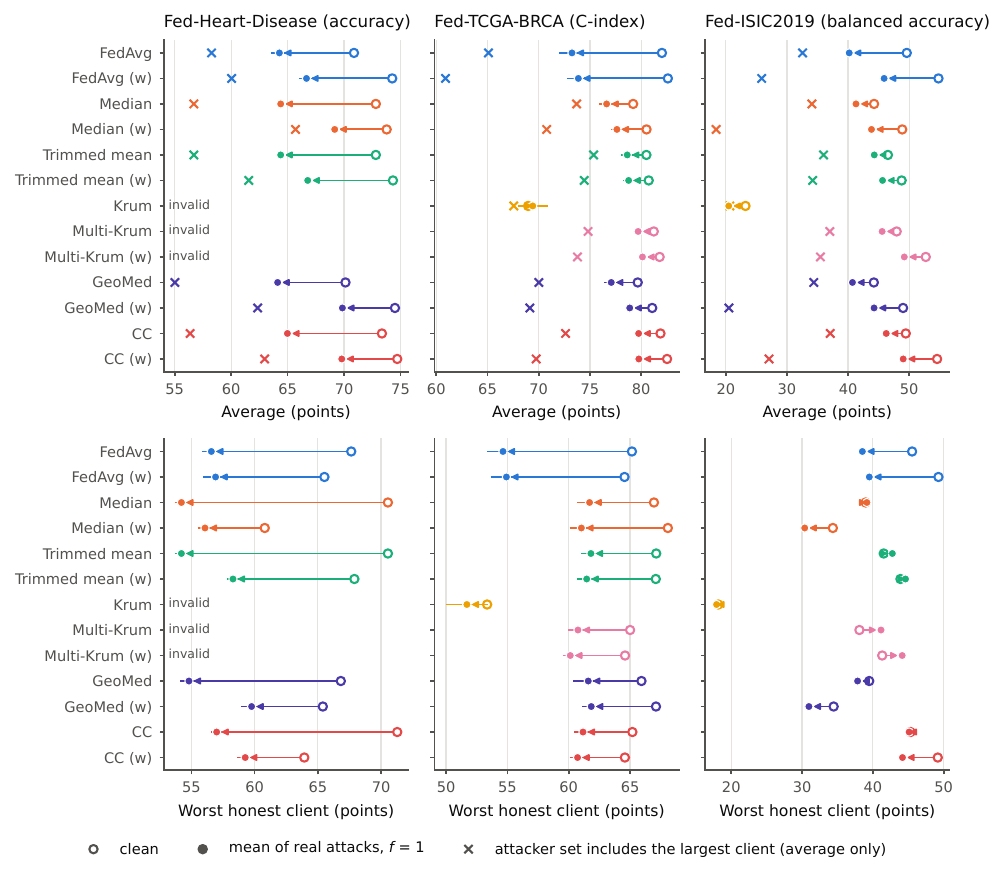}
\caption{One attacker ($f = 1$): average and worst honest client per rule variant, clean (open circles) and under the mean of real attacks (filled), with $\times$ marking the average when the attacker is the largest client.}
\label{fig:f1}
\end{figure*}

When a single hospital turns Byzantine ($f = 1$; Fig.~\ref{fig:f1}), unweighted FedAvg drops by 6.6 points on Heart, 8.8 points on TCGA, and 9.4 points on ISIC (averaged across all real attacks and attacker identities). Robust rules limit this average degradation to 1.5--2.9 points on TCGA and 2.2--5.5 points on ISIC (two attackers: Fig.~\ref{fig:f2} in the Appendix). On Heart, unweighted robust rules fall by 6.0 points (\texttt{geomed}, reaching 64.1\%) to 8.4 points (\texttt{cc}, reaching 65.0\%; \texttt{median}/\texttt{trimmed\_mean}, reaching 64.4\%) compared to 64.3\% for unweighted FedAvg, whereas weighted \texttt{cc}, \texttt{geomed}, and \texttt{median} limit average loss to 4.6--4.9 points (vs.\ 7.6 for weighted FedAvg).

However, pooled averages conceal large damage under stealthy attacks. Our pre-specified \texttt{alie\_z1} attack ($z = 1.0$) hides the malicious update one standard deviation inside the honest coordinate spread. On Fed-Heart-Disease it is the mildest attack on average, leaving unweighted FedAvg's worst honest hospital unchanged ($-0.3 \pm 0.4$ points). Yet because median-type operators track coordinate order statistics, \texttt{alie\_z1} degrades the worst honest hospital by $-9.5 \pm 0.8$ points under median, $-10.0 \pm 0.6$ under median (w), and $-8.6 \pm 0.9$ under geomed.

\subsection{RQ-C: Decomposing honest data loss vs.\ true attack damage ($f = 1$ and $f = 2$)}

\begin{figure*}[!t]
\centering
\includegraphics[width=\textwidth]{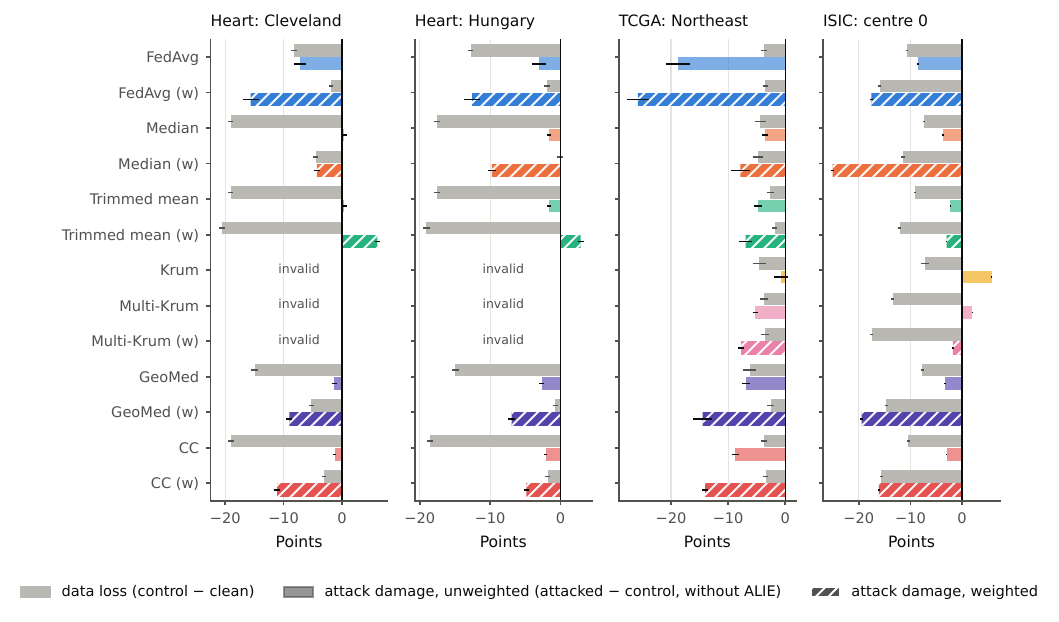}
\caption{Leave-out decomposition when the largest clients attack ($f = 1$): data loss (grey, control $-$ clean) and attack damage without ALIE (colour, attacked $-$ control; hatched = weighted).}
\label{fig:loo_f1}
\end{figure*}

Without leave-out controls, an attack by the largest institution makes every defence appear broken: when centre 0 (53.4\%) attacks on ISIC, unweighted robust rules fall by 10.9 to 13.3 points and weight-dependent rules fall by 31.8 to 36.5 points. Table~\ref{tab:loo} and Fig.~\ref{fig:loo_f1} present the exact-twin leave-out decomposition (reporting true attack damage without ALIE).

\begin{table*}[!t]
\centering
\caption{Leave-out twin decomposition when the largest institution attacks at $f = 1$ (top) and in pairs at $f = 2$ (bottom): honest data loss (control $-$ clean) and attack damage excluding ALIE (attacked $-$ control), in metric points ($\times 100$), mean $\pm$ 95\% CI over seeds.}
\label{tab:loo}
\resizebox{\textwidth}{!}{
\begin{tabular}{@{}l cc cc cc cc@{}}
\toprule
& \multicolumn{2}{c}{Heart: Cleveland} & \multicolumn{2}{c}{Heart: Hungary} & \multicolumn{2}{c}{TCGA: Northeast} & \multicolumn{2}{c}{ISIC: centre 0} \\
\cmidrule(lr){2-3}\cmidrule(lr){4-5}\cmidrule(lr){6-7}\cmidrule(lr){8-9}
Rule & Data loss & Attack & Data loss & Attack & Data loss & Attack & Data loss & Attack \\
\midrule
FedAvg & $-$8.2{\scriptsize\,$\pm$0.5} & $-$7.2{\scriptsize\,$\pm$1.0} & $-$12.8{\scriptsize\,$\pm$0.4} & $-$3.0{\scriptsize\,$\pm$1.0} & $-$3.8{\scriptsize\,$\pm$0.5} & $-$18.8{\scriptsize\,$\pm$2.1} & $-$10.6{\scriptsize\,$\pm$0.2} & $-$8.5{\scriptsize\,$\pm$0.2} \\
FedAvg (w) & $-$1.9{\scriptsize\,$\pm$0.4} & $-$15.7{\scriptsize\,$\pm$1.4} & $-$1.9{\scriptsize\,$\pm$0.4} & $-$12.6{\scriptsize\,$\pm$1.2} & $-$3.5{\scriptsize\,$\pm$0.4} & $-$25.8{\scriptsize\,$\pm$1.9} & $-$15.9{\scriptsize\,$\pm$0.3} & $-$17.6{\scriptsize\,$\pm$0.2} \\
Median & $-$19.1{\scriptsize\,$\pm$0.4} & +0.3{\scriptsize\,$\pm$0.5} & $-$17.6{\scriptsize\,$\pm$0.4} & $-$1.6{\scriptsize\,$\pm$0.3} & $-$4.4{\scriptsize\,$\pm$1.0} & $-$3.6{\scriptsize\,$\pm$0.6} & $-$7.5{\scriptsize\,$\pm$0.2} & $-$3.8{\scriptsize\,$\pm$0.2} \\
Median (w) & $-$4.5{\scriptsize\,$\pm$0.5} & $-$4.3{\scriptsize\,$\pm$0.5} & +0.0{\scriptsize\,$\pm$0.4} & $-$9.8{\scriptsize\,$\pm$0.6} & $-$4.7{\scriptsize\,$\pm$0.9} & $-$7.9{\scriptsize\,$\pm$1.7} & $-$11.4{\scriptsize\,$\pm$0.4} & $-$25.1{\scriptsize\,$\pm$0.3} \\
Trimmed mean & $-$19.1{\scriptsize\,$\pm$0.4} & +0.3{\scriptsize\,$\pm$0.5} & $-$17.6{\scriptsize\,$\pm$0.4} & $-$1.6{\scriptsize\,$\pm$0.3} & $-$2.6{\scriptsize\,$\pm$0.6} & $-$4.7{\scriptsize\,$\pm$0.7} & $-$9.1{\scriptsize\,$\pm$0.2} & $-$2.3{\scriptsize\,$\pm$0.1} \\
Trimmed mean (w) & $-$20.6{\scriptsize\,$\pm$0.5} & +6.0{\scriptsize\,$\pm$0.4} & $-$19.1{\scriptsize\,$\pm$0.5} & +2.9{\scriptsize\,$\pm$0.4} & $-$1.8{\scriptsize\,$\pm$0.4} & $-$6.9{\scriptsize\,$\pm$1.1} & $-$12.1{\scriptsize\,$\pm$0.2} & $-$3.0{\scriptsize\,$\pm$0.1} \\
Krum & -- & -- & -- & -- & $-$4.5{\scriptsize\,$\pm$1.1} & $-$0.8{\scriptsize\,$\pm$1.2} & $-$7.2{\scriptsize\,$\pm$0.7} & +5.7{\scriptsize\,$\pm$0.1} \\
Multi-Krum & -- & -- & -- & -- & $-$3.6{\scriptsize\,$\pm$0.7} & $-$5.2{\scriptsize\,$\pm$0.5} & $-$13.5{\scriptsize\,$\pm$0.2} & +2.0{\scriptsize\,$\pm$0.1} \\
Multi-Krum (w) & -- & -- & -- & -- & $-$3.5{\scriptsize\,$\pm$0.7} & $-$7.7{\scriptsize\,$\pm$0.5} & $-$17.5{\scriptsize\,$\pm$0.2} & $-$1.7{\scriptsize\,$\pm$0.1} \\
GeoMed & $-$15.0{\scriptsize\,$\pm$0.5} & $-$1.4{\scriptsize\,$\pm$0.4} & $-$15.0{\scriptsize\,$\pm$0.5} & $-$2.6{\scriptsize\,$\pm$0.4} & $-$6.2{\scriptsize\,$\pm$1.2} & $-$6.9{\scriptsize\,$\pm$0.7} & $-$7.7{\scriptsize\,$\pm$0.2} & $-$3.3{\scriptsize\,$\pm$0.1} \\
GeoMed (w) & $-$5.3{\scriptsize\,$\pm$0.4} & $-$9.0{\scriptsize\,$\pm$0.5} & $-$0.7{\scriptsize\,$\pm$0.3} & $-$6.9{\scriptsize\,$\pm$0.5} & $-$2.5{\scriptsize\,$\pm$0.6} & $-$14.5{\scriptsize\,$\pm$1.6} & $-$14.6{\scriptsize\,$\pm$0.4} & $-$19.5{\scriptsize\,$\pm$0.3} \\
CC & $-$19.0{\scriptsize\,$\pm$0.5} & $-$1.2{\scriptsize\,$\pm$0.2} & $-$18.5{\scriptsize\,$\pm$0.4} & $-$2.1{\scriptsize\,$\pm$0.2} & $-$3.7{\scriptsize\,$\pm$0.5} & $-$8.7{\scriptsize\,$\pm$0.6} & $-$10.4{\scriptsize\,$\pm$0.2} & $-$2.9{\scriptsize\,$\pm$0.1} \\
CC (w) & $-$3.1{\scriptsize\,$\pm$0.3} & $-$11.1{\scriptsize\,$\pm$0.6} & $-$1.8{\scriptsize\,$\pm$0.4} & $-$4.8{\scriptsize\,$\pm$0.4} & $-$3.5{\scriptsize\,$\pm$0.5} & $-$14.1{\scriptsize\,$\pm$0.5} & $-$15.6{\scriptsize\,$\pm$0.3} & $-$16.1{\scriptsize\,$\pm$0.1} \\
\bottomrule
\end{tabular}
}\\[6pt]
\resizebox{\textwidth}{!}{
\begin{tabular}{@{}l cc cc cc cc@{}}
\toprule
& \multicolumn{2}{c}{TCGA: with Northeast (pairs)} & \multicolumn{2}{c}{TCGA: without Northeast (pairs)} & \multicolumn{2}{c}{ISIC: with centre 0 (pairs)} & \multicolumn{2}{c}{ISIC: without centre 0 (pairs)} \\
\cmidrule(lr){2-3}\cmidrule(lr){4-5}\cmidrule(lr){6-7}\cmidrule(lr){8-9}
Rule & Data loss & Attack & Data loss & Attack & Data loss & Attack & Data loss & Attack \\
\midrule
FedAvg & $-$4.3{\scriptsize\,$\pm$1.6} & $-$22.0{\scriptsize\,$\pm$4.5} & $-$0.4{\scriptsize\,$\pm$0.2} & $-$13.1{\scriptsize\,$\pm$3.8} & $-$13.1{\scriptsize\,$\pm$0.2} & $-$12.0{\scriptsize\,$\pm$0.1} & $-$0.8{\scriptsize\,$\pm$0.2} & $-$15.4{\scriptsize\,$\pm$0.2} \\
FedAvg (w) & $-$4.5{\scriptsize\,$\pm$1.5} & $-$26.3{\scriptsize\,$\pm$2.8} & $-$0.9{\scriptsize\,$\pm$0.5} & $-$11.2{\scriptsize\,$\pm$3.5} & $-$17.9{\scriptsize\,$\pm$0.3} & $-$18.7{\scriptsize\,$\pm$0.2} & $-$1.7{\scriptsize\,$\pm$0.2} & $-$9.2{\scriptsize\,$\pm$0.2} \\
Median & $-$2.7{\scriptsize\,$\pm$1.6} & $-$11.8{\scriptsize\,$\pm$3.6} & $-$0.7{\scriptsize\,$\pm$0.6} & $-$7.2{\scriptsize\,$\pm$1.6} & $-$9.3{\scriptsize\,$\pm$0.1} & $-$6.9{\scriptsize\,$\pm$0.1} & $-$0.3{\scriptsize\,$\pm$0.2} & $-$6.6{\scriptsize\,$\pm$0.2} \\
Median (w) & $-$4.0{\scriptsize\,$\pm$1.7} & $-$16.1{\scriptsize\,$\pm$3.2} & $-$1.5{\scriptsize\,$\pm$1.0} & $-$5.3{\scriptsize\,$\pm$1.6} & $-$16.0{\scriptsize\,$\pm$0.3} & $-$20.6{\scriptsize\,$\pm$0.3} & +0.0{\scriptsize\,$\pm$0.0} & +0.0{\scriptsize\,$\pm$0.1} \\
GeoMed & $-$5.2{\scriptsize\,$\pm$3.4} & $-$6.8{\scriptsize\,$\pm$2.7} & $-$0.6{\scriptsize\,$\pm$1.0} & $-$4.8{\scriptsize\,$\pm$1.2} & $-$9.2{\scriptsize\,$\pm$0.2} & $-$6.1{\scriptsize\,$\pm$0.1} & $-$1.5{\scriptsize\,$\pm$0.2} & $-$6.2{\scriptsize\,$\pm$0.1} \\
GeoMed (w) & $-$6.2{\scriptsize\,$\pm$2.0} & $-$12.0{\scriptsize\,$\pm$3.7} & $-$1.5{\scriptsize\,$\pm$0.4} & $-$1.6{\scriptsize\,$\pm$0.6} & $-$17.2{\scriptsize\,$\pm$0.4} & $-$18.2{\scriptsize\,$\pm$0.2} & $-$0.1{\scriptsize\,$\pm$0.0} & +0.1{\scriptsize\,$\pm$0.1} \\
CC & $-$4.4{\scriptsize\,$\pm$1.7} & $-$10.8{\scriptsize\,$\pm$2.3} & $-$0.5{\scriptsize\,$\pm$0.3} & $-$3.4{\scriptsize\,$\pm$0.6} & $-$12.9{\scriptsize\,$\pm$0.2} & $-$6.2{\scriptsize\,$\pm$0.1} & $-$0.7{\scriptsize\,$\pm$0.2} & $-$6.2{\scriptsize\,$\pm$0.1} \\
CC (w) & $-$4.5{\scriptsize\,$\pm$1.4} & $-$15.0{\scriptsize\,$\pm$1.2} & $-$1.0{\scriptsize\,$\pm$0.6} & $-$2.0{\scriptsize\,$\pm$0.3} & $-$17.6{\scriptsize\,$\pm$0.3} & $-$16.4{\scriptsize\,$\pm$0.1} & $-$1.6{\scriptsize\,$\pm$0.2} & $-$1.8{\scriptsize\,$\pm$0.1} \\
\bottomrule
\end{tabular}
}
\end{table*}

Table~\ref{tab:loo} shows our main empirical finding:
\begin{enumerate}
\item When Cleveland (41\%) or Hungary (35\%) attacks on Fed-Heart-Disease, unweighted median, geomed, and cc suffer $-15.0$ to $-19.1$ points of honest data loss from the hospital's absence, but only $+0.3$ to $-2.6$ points of true attack damage. Similarly, when Centre 0 (53.4\%) attacks on Fed-ISIC2019, unweighted robust rules lose $-7.5$ to $-10.4$ points from missing data while limiting true adversarial damage to just $-2.3$ to $-3.8$ points (and Multi-Krum achieves $+2.0$ points because rejecting Centre 0's poisoned update forces it to include Centre 3, which it otherwise drops in clean runs).
\item Rules where sample weights govern the robust operator (median (w), geomed (w), cc (w)) exhibit small data loss on Heart (0.0 to $-5.3$ points, as the remaining large hospital dominates) but suffer $-4.3$ to $-11.1$ points of true attack damage. On ISIC, where Centre 0 exceeds 50\% weight, true attack damage reaches $-16.1$ (cc (w)), $-19.5$ (geomed (w)), and $-25.1$ points (median (w)), driving worst-honest-centre accuracy down by 31 to 37 points; pooled accuracy under median (w) falls to near-random guessing (11.7--12.6\% under each non-ALIE attack). By contrast, \texttt{trimmed\_mean} (w) and \texttt{multi\_krum} (w), which trim or select candidates strictly by client count before weighting, behave like unweighted rules and restrict Centre 0's attack damage to $-3.0$ and $-1.7$ points.
\item On Fed-TCGA-BRCA, Northeast holds 28.6\% of training patients, well below the 50\% dictatorship threshold. Because data loss is small ($-1.8$ to $-6.2$ points), most degradation is true attack damage. Even here, sample-size weighting amplifies Northeast's attack damage by $+2.2$ to $+7.6$ points across every robust rule at $f = 1$ (e.g., GeoMed worsens from $-6.9$ unweighted to $-14.5$ weighted; CC from $-8.7$ to $-14.1$). At $f = 2$ (pairs including Northeast, holding 33--48\% combined share; Fig.~\ref{fig:loo_f2} in the Appendix), weighted rules suffer 1.4$\times$ to 1.8$\times$ the attack damage of unweighted rules.
\end{enumerate}

\subsection{Why hospital participation still matters (local test set audit)}

Finally, evaluating our clean leave-one-out controls on the excluded hospital's own test set quantifies the clinical penalty of opting out of a federation. Staying out of the federation costs a hospital up to 34.4 percentage points of accuracy on its own patients on Heart (Hungary under weighted Trimmed Mean), up to 14.9 C-index points on TCGA (Europe), and up to 42.0 points on ISIC. On TCGA, South shows a $+2$ to $+17$ point anomaly for 11 of 13 rule variants when left out; South has only 4 test deaths (44 pairs), so we do not interpret this. Another exception occurs under ISIC's weighted median and GeoMed: because Centre 0 acts as a dictator, centres 1 through 5 lose 0.0 to 0.7 points when leaving the federation, confirming that under a $> 50\%$ weighted median, minority hospitals gain nothing from participating.

\section{Discussion and practical recommendations}

\subsection{Deployment checklist for small clinical consortia}

Our findings translate into four concrete engineering rules for deploying Byzantine-robust FL across hospital networks:
\begin{enumerate}
\item Audit combinatorial validity before selecting an aggregator. Never deploy Bulyan in consortia with $n \leq 6$ hospitals, and never deploy Krum or Multi-Krum at $n = 4$ ($f \geq 1$) or $n = 6$ ($f \geq 2$).
\item Keep sample weights out of the robust filtering step. Whenever a single hospital holds $> 50\%$ of records (or an $f = 2$ pair can exceed 50\%), sample-weighted median, geometric median, and centered clipping must not be used. Even at 29\% share (TCGA), weighting inside the robust operator multiplies adversarial attack damage by 1.6 to 2.2 ($f = 1$). If sample weighting is desired, apply weights only after count-based selection (\texttt{trimmed\_mean} (w) or \texttt{multi\_krum} (w)), or use unweighted Centered Clipping (\texttt{cc}), which provided the best overall trade-off across all three datasets (near-zero clean cost and $\leq 2.9$ points of attack damage when the largest hospital attacked on Heart and ISIC).
\item Use leave-out controls when interpreting stress tests. In small federations, losing a major hospital's honest gradients  lowers accuracy. Without leave-out controls, practitioners risk discarding working count-based defences by mistaking honest data loss for adversarial failure.
\item Monitor per-hospital validation metrics. Pooled averages fail to detect clean dictatorships (ISIC) and stealthy in-distribution attacks (\texttt{alie\_z1} on Heart) that degrade the worst-served hospital.
\end{enumerate}

\subsection{Limitations and future work}

First, our audit covers three clinical datasets with $n \in \{4, 6\}$ institutions using linear and logistic models (and a linear probe on frozen EfficientNet-B0 features for ISIC to make its 27,620 main-grid runs feasible). While frozen features yield lower absolute accuracy than full fine-tuning, every aggregation rule observes identical representations. Second, TCGA test splits contain only 32 total patient deaths (1 in Canada, 2 in Europe), making regional TCGA fairness metrics noisier than ISIC's. For example, on Fed-TCGA-BRCA ($f = 1$), \texttt{alie\_z1} raises the stratified C-index by $+1.0$ to $+2.9$ points for every rule except Krum ($+7.4$) while the worst honest region drops by 2.9 to 8.8 points; under median, South loses 14 to 17 points whenever Northeast, Midwest, Europe, or Canada attacks, and Northeast alone holds 59\% of the within-region test pairs. Because South has only 4 test deaths, we report this as a remark, not a finding. Third, we assume hospitals report true sample sizes $m_i$; allowing Byzantine hospitals to inflate $m_i$ would further accelerate the breakdown of weighted rules. Future work should extend exact-twin leave-out auditing to parameter-efficient fine-tuning (PEFT/QLoRA) of local clinical Large Language Models, where low-rank adapter factors ($B \cdot A$) interact non-linearly during coordinate-wise aggregation.

\section*{Code Availability}
All code, the pre-specified protocol and the scripts that rebuild every table and figure are available at \url{https://github.com/adnanmalkani/byzantine-fl-bench} (release v1.0-manuscript).

\section*{Acknowledgments}
Parts of the experiment and analysis code were developed with the assistance of an AI coding tool (Claude Code, Anthropic). All code was reviewed, tested and validated against published baselines by the author.

\bibliographystyle{IEEEtran}
\bibliography{references}

\appendices

\section{Additional figures and tables}
Figs.~5--7 and Table~VI report the two-attacker results, the two-attacker leave-out decomposition, the FLamby reproductions and the $k = 10$ sensitivity check.

\begin{figure*}[!t]
\centering
\includegraphics[width=0.7\textwidth]{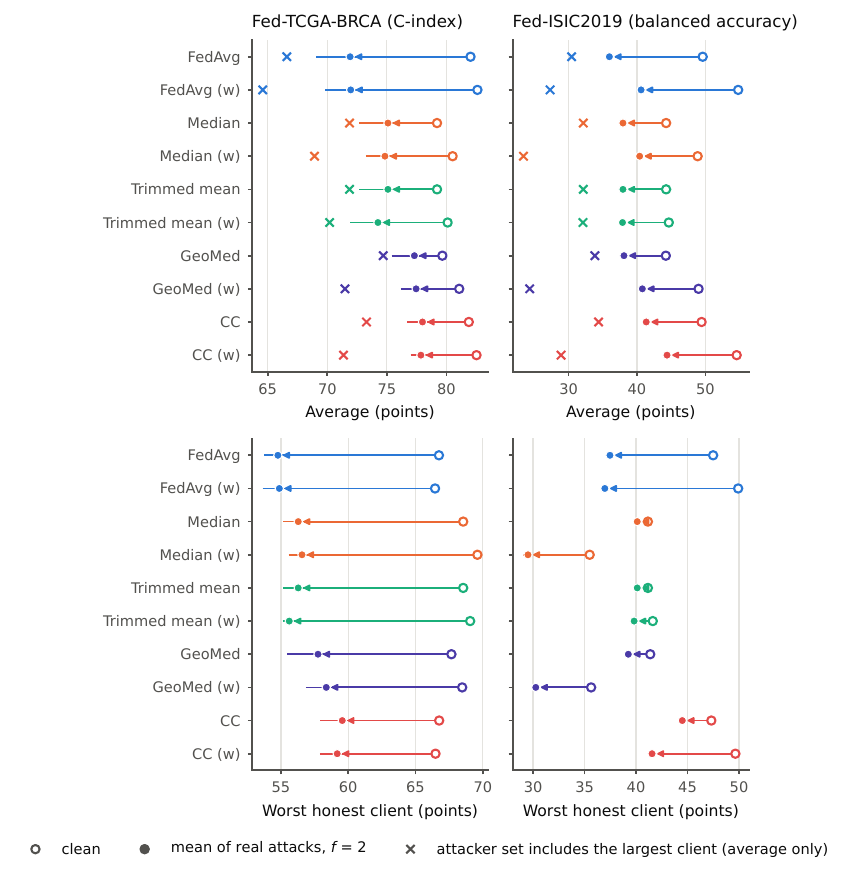}
\caption{Two attackers ($f = 2$) on Fed-TCGA-BRCA and Fed-ISIC2019, drawn as in Fig.~\ref{fig:f1} (on Fed-Heart-Disease only FedAvg is valid at $f = 2$).}
\label{fig:f2}
\end{figure*}

\begin{figure*}[!t]
\centering
\includegraphics[width=\textwidth]{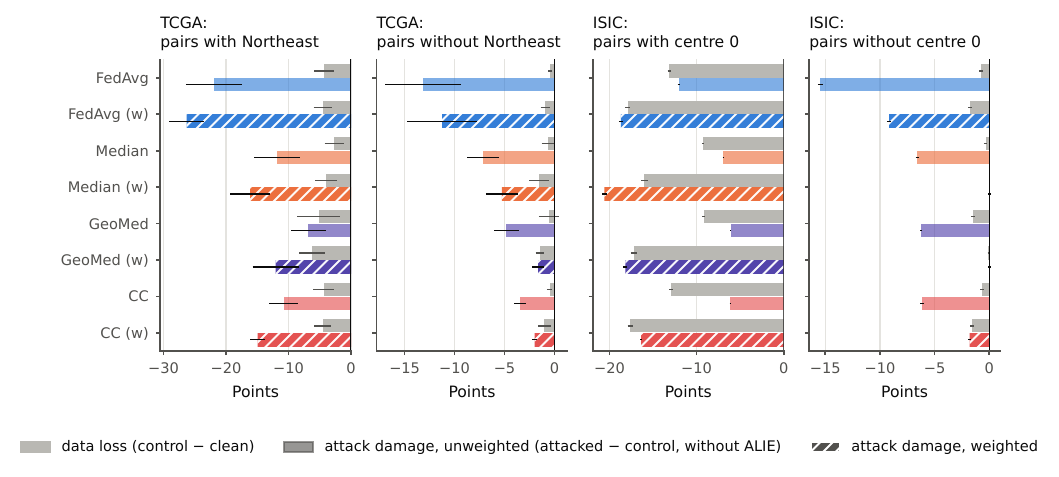}
\caption{Leave-out decomposition at $f = 2$, averaged over the attacker pairs with and without the largest client (Fed-TCGA-BRCA and Fed-ISIC2019).}
\label{fig:loo_f2}
\end{figure*}

\begin{figure*}[!t]
\centering
\includegraphics[width=\textwidth]{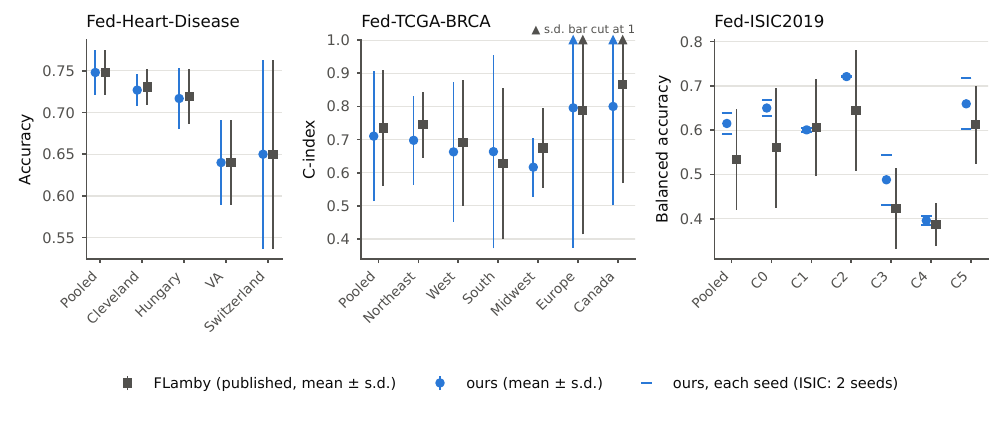}
\caption{FLamby reproductions: our FedAvg against FLamby's published mean $\pm$ s.d.\ on every test set (the TCGA axis is capped at 1 and bars cut there are marked $\blacktriangle$).}
\label{fig:repro}
\end{figure*}

\begin{table*}[!t]
\centering
\caption{Sensitivity of the Fed-ISIC2019 worst honest centre to the class threshold: balanced accuracy over the classes with at least $k = 5$ (main text) or $k = 10$ test images at that centre (centre 5 drops out at $k = 10$), clean (scored over the same honest centres) and under the mean of real attacks, at $f = 1$ and $f = 2$ (mean $\pm$ 95\% CI over seeds).}
\label{tab:k10}
\resizebox{\textwidth}{!}{
\begin{tabular}{@{}l cccc cccc@{}}
\toprule
& \multicolumn{4}{c}{$f=1$} & \multicolumn{4}{c}{$f=2$} \\
\cmidrule(lr){2-5}\cmidrule(l){6-9}
& \multicolumn{2}{c}{$k=5$} & \multicolumn{2}{c}{$k=10$} & \multicolumn{2}{c}{$k=5$} & \multicolumn{2}{c}{$k=10$} \\
\cmidrule(lr){2-3}\cmidrule(lr){4-5}\cmidrule(lr){6-7}\cmidrule(l){8-9}
Rule & Clean & Attacked & Clean & Attacked & Clean & Attacked & Clean & Attacked \\
\midrule
FedAvg & 45.5{\scriptsize\,$\pm$0.2} & 38.5{\scriptsize\,$\pm$0.2} & 45.5{\scriptsize\,$\pm$0.2} & 38.2{\scriptsize\,$\pm$0.2} & 47.5{\scriptsize\,$\pm$0.2} & 37.5{\scriptsize\,$\pm$0.1} & 47.5{\scriptsize\,$\pm$0.2} & 37.2{\scriptsize\,$\pm$0.1} \\
FedAvg (w) & 49.3{\scriptsize\,$\pm$0.3} & 39.5{\scriptsize\,$\pm$0.3} & 49.3{\scriptsize\,$\pm$0.3} & 41.0{\scriptsize\,$\pm$0.3} & 49.9{\scriptsize\,$\pm$0.3} & 37.0{\scriptsize\,$\pm$0.2} & 49.9{\scriptsize\,$\pm$0.3} & 37.7{\scriptsize\,$\pm$0.2} \\
Median & 39.0{\scriptsize\,$\pm$0.2} & 39.2{\scriptsize\,$\pm$0.1} & 39.0{\scriptsize\,$\pm$0.2} & 39.2{\scriptsize\,$\pm$0.1} & 41.2{\scriptsize\,$\pm$0.2} & 40.1{\scriptsize\,$\pm$0.1} & 41.2{\scriptsize\,$\pm$0.2} & 40.1{\scriptsize\,$\pm$0.1} \\
Median (w) & 34.3{\scriptsize\,$\pm$0.8} & 30.4{\scriptsize\,$\pm$0.6} & 34.1{\scriptsize\,$\pm$0.8} & 30.2{\scriptsize\,$\pm$0.7} & 35.5{\scriptsize\,$\pm$0.7} & 29.5{\scriptsize\,$\pm$0.4} & 34.9{\scriptsize\,$\pm$0.8} & 29.1{\scriptsize\,$\pm$0.5} \\
Trimmed mean & 41.5{\scriptsize\,$\pm$0.2} & 42.7{\scriptsize\,$\pm$0.1} & 41.5{\scriptsize\,$\pm$0.2} & 42.7{\scriptsize\,$\pm$0.1} & 41.2{\scriptsize\,$\pm$0.2} & 40.1{\scriptsize\,$\pm$0.1} & 41.2{\scriptsize\,$\pm$0.2} & 40.1{\scriptsize\,$\pm$0.1} \\
Trimmed mean (w) & 43.9{\scriptsize\,$\pm$0.2} & 44.6{\scriptsize\,$\pm$0.1} & 43.9{\scriptsize\,$\pm$0.2} & 44.6{\scriptsize\,$\pm$0.1} & 41.6{\scriptsize\,$\pm$0.2} & 39.8{\scriptsize\,$\pm$0.1} & 41.6{\scriptsize\,$\pm$0.2} & 39.8{\scriptsize\,$\pm$0.1} \\
Krum & 18.2{\scriptsize\,$\pm$0.1} & 17.9{\scriptsize\,$\pm$0.1} & 18.2{\scriptsize\,$\pm$0.1} & 18.0{\scriptsize\,$\pm$0.1} & \multicolumn{4}{c}{invalid} \\
Multi-Krum & 38.1{\scriptsize\,$\pm$0.3} & 41.2{\scriptsize\,$\pm$0.1} & 38.1{\scriptsize\,$\pm$0.3} & 41.2{\scriptsize\,$\pm$0.1} & \multicolumn{4}{c}{invalid} \\
Multi-Krum (w) & 41.3{\scriptsize\,$\pm$0.5} & 44.1{\scriptsize\,$\pm$0.3} & 41.3{\scriptsize\,$\pm$0.5} & 44.2{\scriptsize\,$\pm$0.3} & \multicolumn{4}{c}{invalid} \\
GeoMed & 39.5{\scriptsize\,$\pm$0.2} & 37.8{\scriptsize\,$\pm$0.1} & 39.5{\scriptsize\,$\pm$0.2} & 37.8{\scriptsize\,$\pm$0.1} & 41.4{\scriptsize\,$\pm$0.1} & 39.3{\scriptsize\,$\pm$0.1} & 41.4{\scriptsize\,$\pm$0.1} & 39.3{\scriptsize\,$\pm$0.1} \\
GeoMed (w) & 34.5{\scriptsize\,$\pm$0.8} & 31.0{\scriptsize\,$\pm$0.6} & 34.2{\scriptsize\,$\pm$0.8} & 30.7{\scriptsize\,$\pm$0.6} & 35.6{\scriptsize\,$\pm$0.6} & 30.3{\scriptsize\,$\pm$0.4} & 35.1{\scriptsize\,$\pm$0.7} & 29.9{\scriptsize\,$\pm$0.5} \\
CC & 45.3{\scriptsize\,$\pm$0.2} & 45.1{\scriptsize\,$\pm$0.1} & 45.3{\scriptsize\,$\pm$0.2} & 45.1{\scriptsize\,$\pm$0.1} & 47.3{\scriptsize\,$\pm$0.2} & 44.5{\scriptsize\,$\pm$0.1} & 47.3{\scriptsize\,$\pm$0.2} & 44.5{\scriptsize\,$\pm$0.1} \\
CC (w) & 49.2{\scriptsize\,$\pm$0.3} & 44.2{\scriptsize\,$\pm$0.2} & 49.2{\scriptsize\,$\pm$0.3} & 44.8{\scriptsize\,$\pm$0.2} & 49.6{\scriptsize\,$\pm$0.3} & 41.6{\scriptsize\,$\pm$0.1} & 49.6{\scriptsize\,$\pm$0.3} & 42.5{\scriptsize\,$\pm$0.1} \\
\bottomrule
\end{tabular}
}
\end{table*}

\end{document}